\documentclass{article}

\usepackage[preprint]{neurips_2024}

\usepackage{natbib} 
\usepackage[utf8]{inputenc}
\usepackage[colorlinks=true, citecolor=blue]{hyperref} 

\usepackage[utf8]{inputenc} 
\usepackage[T1]{fontenc}    
\usepackage{url}            
\usepackage{booktabs}       
\usepackage{subcaption} 
\usepackage{amsfonts}       
\usepackage{nicefrac}       
\usepackage{microtype}      
\usepackage{xcolor}         
\usepackage{graphicx} 
\usepackage{chngcntr}
\usepackage{amsmath} 
\usepackage{wrapfig}
\usepackage{array} 
\usepackage{listings} 
\usepackage{placeins} 

\lstdefinestyle{mystyle}{
    backgroundcolor=\color{white},   
    commentstyle=\color{black},
    keywordstyle=\color{blue},
    numberstyle=\tiny\color{gray},
    stringstyle=\color{purple},
    basicstyle=\ttfamily\footnotesize,
    breakatwhitespace=false,         
    breaklines=true,                 
    captionpos=b,                    
    keepspaces=true,                 
    numbers=left,                    
    numbersep=5pt,                  
    showspaces=false,                
    showstringspaces=false,
    showtabs=false,                  
    tabsize=2
}

\title{Towards Transparency: Exploring LLM Trainings Datasets through Visual Topic Modeling and Semantic Frame}

\author{%
  Charles de Dampierre\thanks{https://charlesdedampierre.github.io/} \\
  Institut Jean Nicod, Département d’études cognitives\\
  ENS, EHESS, PSL University, CNRS\\
  \texttt{charlesdedampierre@gmail.com} \\
  \And
Andrei Mogoutov\\
 SciencePo Medialab\\
  \texttt{mogoutov@gmail.com} \\
  \AND
 Nicolas Baumard
\thanks{https://nicolasbaumards.org/} \\
  Institut Jean Nicod, Département d’études cognitives\\
  ENS, EHESS, PSL University, CNRS\\
  \texttt{nbaumard@gmail.com
} \\
}

\begin{document}

\maketitle

\begin{abstract}
LLMs are now responsible for taking many decisions on behalf of humans: from answering questions to classifying things, they have become an important part of everyday life. While computation and model architecture have been rapidly expanding in recent years, the efforts towards curating training datasets are still at their beginnings. This underappreciation of training datasets has led LLMs to create biased and low-quality content. In order to solve that issue, we present Bunka, a software that leverages AI and Cognitive Science to improve the refinement of textual datasets. We show how Topic Modeling coupled with 2-dimensional Cartography can increase the transparency of datasets. We then show how the same topic modeling techniques can be applied to Preferences datasets to accelerate the fine-tuning process and increase the capacities of the model on different benchmarks. Lastly, we show how using Frame Analysis can give insights on the existing bias in the training corpus. Overall, we argue that we need better tools to explore and increase the quality and transparency of LLMs training datasets.

\end{abstract}

\section{Introduction}

Information has become a highly demanded commodity in the 21st century. Large Language Models (LLMs) like BERT \citep{devlin_bert_2019}, GPT-3 \citep{brown_language_2020}, and Mixtral \citep{jiang_mixtral_2024} process a lot of data for their training through Transformers \citep{vaswani_attention_2023} and Mamba \citep{gu_mamba_2023} architecture. Thanks to it, they have reached human-like levels in benchmarks related to language, reasoning \citep{huang_towards_2023}, mathematics \citep{azerbayev_llemma_2023}, coding \citep{li_starcoder_2023}, medicine \citep{omiye_large_2023} and other sets of quantitative and qualitative tasks \citep{srivastava_beyond_2023, moro_large_2023}. As of today, the conventional approach to enhance the capabilities of these models is to scale both the datasets and the computational power. Consequently, LLMs have been relying on massive corpora like CommonCrawls, Colossal Clean Crawled Corpus (C4) \citep{zhu_multimodal_2023} or The Pile \citep{gao_pile_2020}. For instance, the latter contains 825 GiB of text, 38\% of which is academic material, 15\% books, 16.6\% social media, 3.1\% language texts, 18.1\% webpages, 7.6\% code and 1.5\% encyclopedia \citep{liu_datasets_2024}. Because of an early focus on size and scaling \citep{hoffmann_training_2022}, first models are trained using those big corpora with too little focus on the quality or explainability of their different sections given their size.  The lack of efficient tools for data pre-processing, data cleaning and data explainability,  led to various issues: inadvertent collection of private data \citep{bae_security_2021} and hateful content \citep{zhou_making_2023}; a predominance of English-language information and biases against less popular languages; the computational costs of training when using huge amount of data \citep{touvron_llama_2023} and its impact on the environment \citep{rillig_risks_2023} and a general lack of transparency in AI \citep{liao_ai_2023}.

The first “patch-methods” to this issue was simply focused on using long system prompts (as in Claude, ChatGPT) to align the LLM with human interests hence banning any negativity and disclosing of private data. Other methods imply retraining very small models in the context of the  BabyLM Challenge \citep{oba_babylm_2023}. More recently, there has been a common effort to enhance the quality of pre-training datasets driven by the consensus that better data leads to better models \citep{gunasekar_textbooks_2023}. Models like Yi \citep{ai_yi_2024} are trained on meticulously refined data representing 0.75\% the size of GPT 4 training datasets. During its pre-training phase, Yi has processed 6B tokens with 8 passes (50B tokens processed overall), and reached an accuracy of 50.6\% on HumanEval and 55.5\% on Mostly Basic Python Programming (MBPP) and the phi-1-small model (350M parameters) achieves 45\% on HumanEval. This has been possible thanks to a variety of new tools to filter large pre-training corpora: perplexity analysis \citep{meister_language_2021}, URL filtering, deduplication \citep{lee_deduplicating_2022}, toxicity detection \citep{zhang_efficient_2023}, privacy content detection and license infringement detection \cite{li_digger_2024} leading to the compilation of refined corpus such as RefinedWeb \citep{penedo_refinedweb_2023}, RedPajama, CC-Stories, and RealNews. Other methods like fine-tuning have focused on continuous training to steadily remove any undesired behaviors or specialize the model to a specific task. In order to reach a specific dataset size, some projects have relied on crowdsourcing like OpenAssistant Conversations \citep{kopf_openassistant_2023}. Others have relied on using synthetic data, i.e. data created by LLM themselves to enhance data quality and data availability \citep{villalobos_will_2022}: Microsoft Phi1 is the first model to have been fine-tuned on 1B tokens of Python textbooks generated by GPT-3.5 \citep{gunasekar_textbooks_2023}.

Despite those advances, there remains transparency issues with the pre-training and fine-training data. Optimatilly, AI creators should be aware of every piece of information that the model ingests but this process takes too much time. The first solution to that issue is to categorize the content in categories like \textit{books}, \textit{social media} or \textit{webpages}, \textit{ArXiv}, \textit{PubMed}, but those are still too broad and are created \textit{a priori}, sometimes not reflecting accurately enough the content of the datasets they describe. For instance, the French Books database (Gallica) uses a system created in the 19th century to classify books (the Dewey system). Another issue is that crowdsourcing data and synthetic data still needs post-processing to remove mistakes, hallucinations or potential bias either due to respondents answers or due to the past training of LLMs.

Data exploration by experts still remains necessary to achieve high quality training dataset and create models which later actions can be explained. It looks as if the lack of focus on training data exploration was due to a lack of tools and methodologies available to do so as \cite{gunasekar_textbooks_2023} put it: "\textit{One challenge is to ensure that the dataset covers all the relevant content and concepts that one wants the model to learn, and that it does so in a balanced and representative way. We lack a good methodology to measure and evaluate the amount of diversity and redundancy in the data}".

\subsection{Related works}

Different methods and frameworks have appeared to increase the transparency of AI datasets such as creating better Data Cards \citep{pushkarna_data_2022}, better dataset documentation \citep{rostamzadeh_healthsheet_2022} or implementing data governance \citep{piktus_roots_2023} and the need for better Human Computer Interaction (HCI) \citep{liao_ai_2023}. Recently, the field of Human Computer Interaction (HCI) has gained importance in dataset quality assessment \citep{liao_ai_2023}. HCI facilitates human evaluation of datasets, particularly useful for textual content, which often contains subtle and implicit meanings that necessitate human involvement which is especially critical when the dataset's content requires expertise that only a few individuals possess \citep{liao_ai_2023, holland_dataset_2018, arnold_factsheets_2019}. In order to further explore this problem, we introduce new solutions built in the BunkaTopics package, to refine training mid-sized datasets and make LLMs more explainable. We describe three use cases: the first use case aims at visually summarizing the prompts of a fine-tuning dataset. The second dataset aims at using Topic Modeling to refine a Reinforcement Learning dataset. The third use case shows how to use Semantic Frames to explore different biases in a dataset.

\section{Use Case 1: Using Topic Cartography to summarize Prompt}
\subsection{Framework: Topic Modeling Cartography}

Textual datasets are complex entities that require time and resources to be easily understood by AI creators. Two complementary approaches can be used to make better sense of them: Topic Modeling and Cartography. Topic Modeling is an old technique in Natural Language Processing (NLP) that has been recently leveraged to filter datasets prior to training \citep{ai_yi_2024}. It consists of finding limited \textit{themes} or \textit{topics} in the data (as opposed to categories designed \textit{a priori}). First approaches were using statistical distribution of words in documents such Latent Dirichlet Allocation (LDA) \citep{blei_latent_nodate} and Non-Negative Matrix Factorization (NMF) \citep{fevotte_algorithms_2011}. Then, word embeddings techniques such as Word2Vec \citep{mikolov_efficient_2013} and Doc2Vec \citep{le_distributed_2014} leveraged the fact that distance between embeddings has been shown to correlate with human ratings of similarity \citep{mikolov_efficient_2013,pennington_glove_2014} to create clusters of words. More recently, encoding-decoding architectures like BERT \citep{devlin_bert_2019} and RoBERTa \citep{liu_roberta_2019} are fine-tuned on Sentence Tasks Similarity (STS) tasks \citep{reimers_sentence-bert_2019} to create more efficient topic-modeling approach like Top2Vec \citep{angelov_top2vec_2020} and BERTopic \citep{grootendorst_bertopic_2022}. Yet, the relationship between the global perspective (topics) and the local perspective (documents), as well as the relationships between topics themselves are still an issue. Cognitive science shows that 2D maps and diagrams are the easiest way to represent the multiple dimensions of an information (distribution of topics, relationships between documents etc.) in a cognitively tractable way \citep{olshannikova_visualizing_2015, harold_cognitive_2016}. Recent advances in neurosciences suggest that the human brain uses the same neuronal resources to represent physical and abstract spaces \citep{bellmund_navigating_2018}. These abstract spaces appear to rely on the same cells, the so-called place and grid-cells of the hippocampus, that are used to encode spatial information \citep{okeefe_hippocampus_1971}. This intuition is present in Tobler’s first law of geography implying that \textit{everything is related to everything else, but near things are more related than distant things}. Furthermore, concepts have an internal coherence and their \textit{'quality dimensions'} are derived from perceptual mechanisms: to some extent, concepts can be represented visually \citep{gardenfors_conceptual_2004}. The objective of Information Cartography is to display textual datasets as a map to leverage the ability of the human brain to associate distances, similarity and meaning \citep{hografer_state_2020}. In this regard \cite{roux_using_2016} noted the extensive list of mapping-related tools: ExploViz, PATHS project \citep{agirre_paths_2013}, reference map \citep{nocaj_organizing_2012}, LDAvis \citep{sievert_ldavis_2014} etc.  Recents advances in the field of non-linear dimension reductions like UMAP \citep{mcinnes_umap_2020} or TSNE \citep{cai_theoretical_2022} have accelerated the development of new 2D visualization leveraging embeddings with the development of tools such as Wizmap \citep{wang_wizmap_2023} and Nomic Atlas.

\begin{figure}[h] %
  \centering 
  \includegraphics[width=12cm]{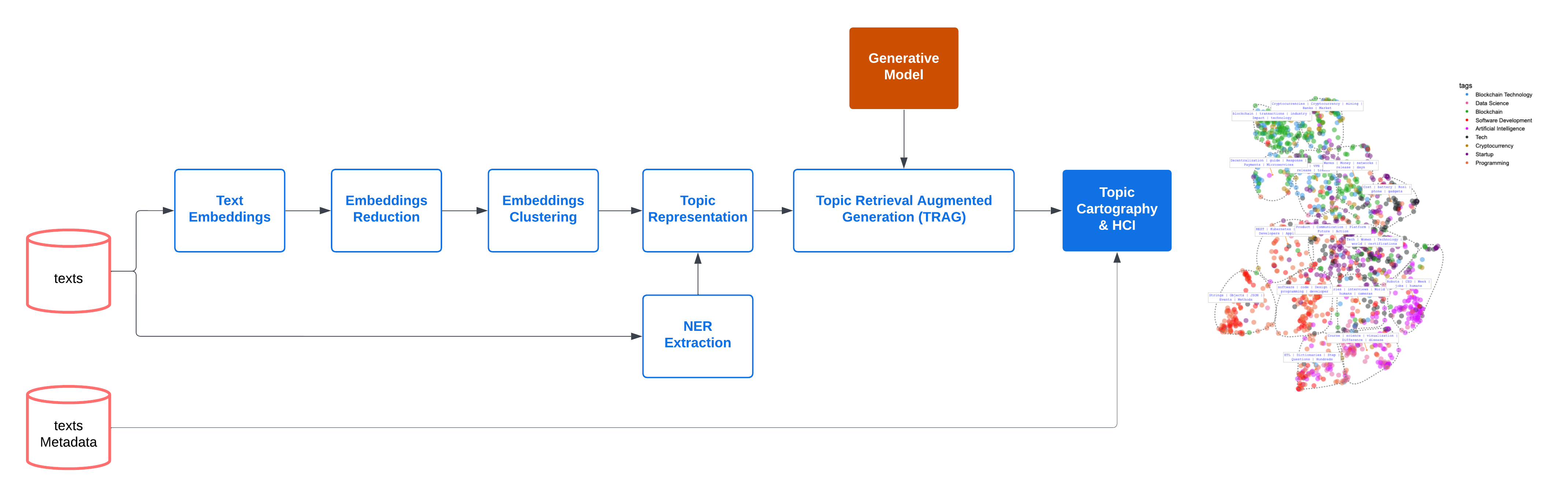} 
  \caption{BunkaTopics Architecture} 
  \label{fig:bunka-architecture}
\end{figure}

\textbf{Infrastructure of BunkaTopics} - BunkaTopics is a python package that leverages Topic Modeling and Human Computer Interaction (HCI) to make sense of large corpus. The software takes a list of textual content as an input and outputs a 2-dimensional map. In between, it can use different Embedding architectures such as SentenceTransformers \citep{reimers_sentence-bert_2019} or FlagEmbedding \citep{chen_bge_2024} to transform the textual content in a latent space. Similar to BERTopic \citep{grootendorst_bertopic_2022}, various techniques for dimension reduction can be chosen (such as UMAP or TSNE), and different clustering methods can then be applied (KMeans, DBSCAN) (see Figure~\ref{fig:bunka-architecture}).

\textbf{Topic Representation} - We then extract Nouns using the SpaCy-based Textacy package and only keep the top 10\% overall nouns to avoid low-quality nouns. We then name the clusters with the 10 most specific nouns using Chi2 metrics \citep{grootendorst_bertopic_2022}. It is possible to either manually change the name of the clusters based on this noun-based label or prompt a LLM to do so. BunkaTopics locates the clusters' name on the map at their centroid and uses the Convex Hull algorithm \citep{chazelle_optimal_1993} to draw limitations around the clusters and Kernel Density Estimation \citep{rosenblatt_central_1956} to indicate the density of documents in the map (the bluer, the denser, see Figure~\ref{fig:bunka-map}). 

\textbf{Ranking Documents} - For every cluster, Bunka ranks the documents it contains based on the number of cluster-specific nouns they contain. For instance, if a document contains 5 specific nouns out of the first 20 specific nouns of the cluster, it is likely to appear first on the navigation bar on the right panel. We then display the results through a React \& D3.js front-end. In an alternative front-end, metadata can be added to the visualization and colored in the final map to highlight specific dimensions within the data (see Figure~\ref{fig:bunka-architecture}).

\subsection{Method}

To examine community-generated prompts that are used for model fine-tuning, we selected the prompt-collective dataset (N=9,333).  We embedded this dataset using the \textbf{mxbai-embed-large-v1} model, the best model with less than 350 millions parameters on the  Massive Text Embedding Benchmark (MTEB) benchmarks \citep{muennighoff_mteb_2023} as of May 2024. We then used UMAP to create coordinates on 2 dimensional-spaces (now called the Map). We applied KMeans clustering and manually set the number of clusters to 15 to achieve an optimal balance between the granularity of analysis and ease of visualization, as too many clusters can complicate the readability of the Map. Next, we extracted nouns (unigrams and bigrams) from the dataset with Part-of-Speech (POS) recognition from the Textacy package. We labeled each cluster by selecting the 10 most specific nouns per cluster, utilizing Chi2 statistics as described by \cite{grootendorst_bertopic_2022}. We selected n=10 for the number of nouns per cluster to quickly understand the meaning of the clusters without overwhelming details given the fact that the nouns are ordered by order of specificity (the first noun being the most specific noun of the cluster). When a unigram was contained inside another bigram in the topic name, we removed the unigram as we find that bigrams often encapsulate more information.

\begin{figure}[h] %
  \centering 
  \includegraphics[width=14cm]{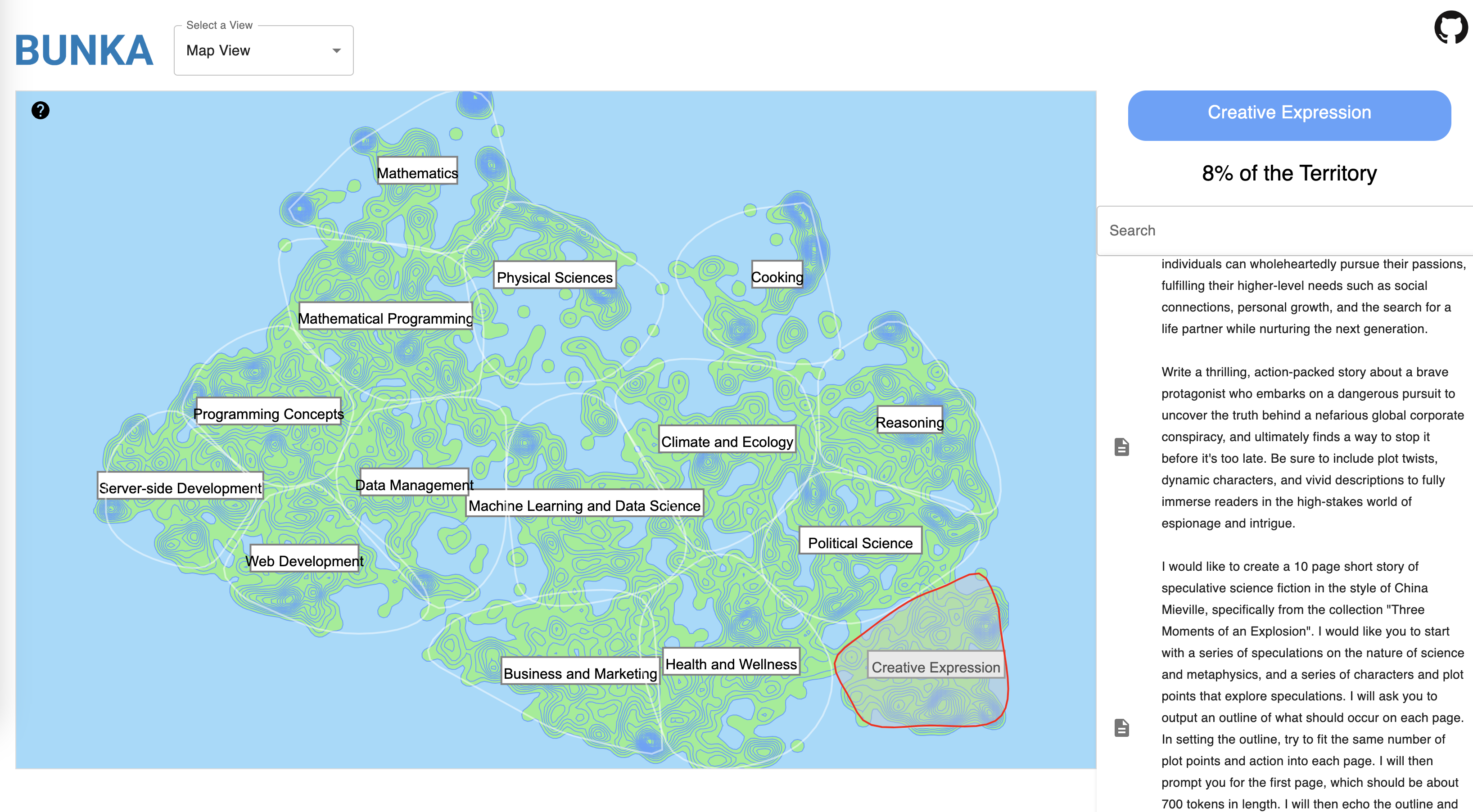} 
  \caption{Bunka Map of the prompt-collective dataset using the \textbf{mxbai-embed-large-v1 model}} 
  \label{fig:bunka-map}
\end{figure}

Results show different topics associated with \textit{Web Development}, \textit{Mathematics}, \textit{Business and Marketing}, \textit{Creative Expression} or \textit{Physical Sciences}. Some conclusions can be made regarding the distance between topics: the similarity between \textit{Web development} \& \textit{Business marketing}, compared to \textit{Mathematics} highlights the differences between application and theory while the similarity between \textit{Cooking} and \textit{Physical Sciences} highlights the common use of materials-related terms. \textit{Psychology} \& \textit{Political Science}, both dealing with human behaviors are close in the latent space (see Figure~\ref{fig:bunka-map}).

\begin{figure}[h] %
  \centering 
  \includegraphics[width=6cm]{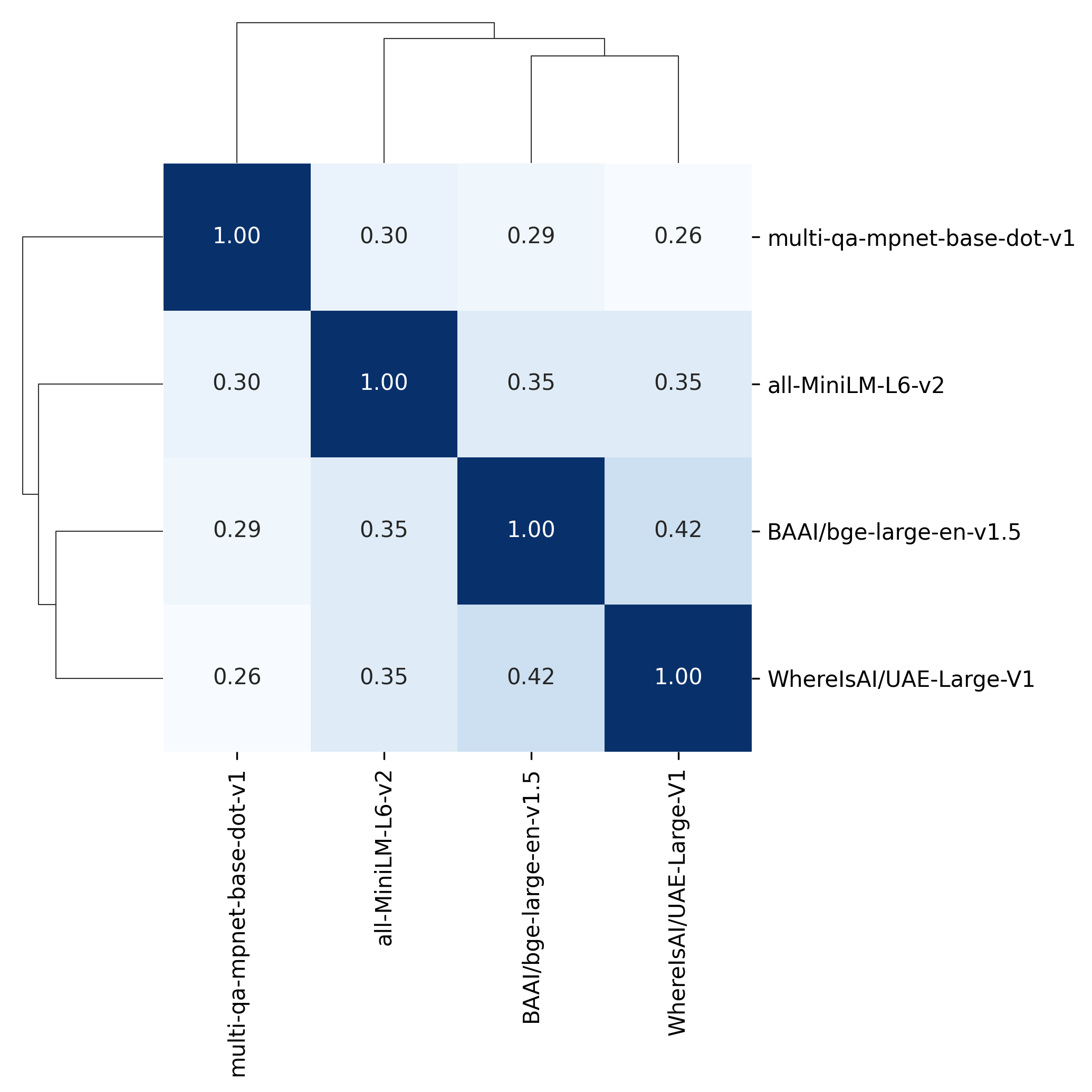} 
  \caption{Adjusted Rand Index (ARI) between the clustered documents from 4 different embedding models. 1 means high similarity.
  } 
  \label{fig:bunka-ari}
\end{figure}

We then perform Topic Modeling (with 15 fixed clusters) using 4 other embedding models with high scores and low number of parameters on the MTEB Leaderboard (\textbf{all-MiniLM-L6-v2}, \textbf{bge-large-en-v1.5}, \textbf{multi-qa-mpnet-base-dot-v1}, \textbf{UAE-Large-V1}) and display the 4 maps (see Figure~\ref{fig:bunka-ari}). In order to quantify the difference, we used the Adjusted Rand Index (ARI) to compare how documents cluster together: a high ARI means that 2 embedding models make documents clustered in a similar way. While the overall shape varies, the results suggest that the clusters remain fairly similar: we find that the 2 best ranked models on MTEB Leaderboard (\textbf{ge-large-en-v1.5} and \textbf{UAE-Large-V1}) create similar topics (ARI=0.42) (see Figure~\ref{fig:bunka-ari}). The Maps created can be found in Annexe~\ref{annexe:bunka-maps-examples}.

\FloatBarrier 

\section{Use Case 2: Topic Modeling to clean datasets for Direct Preference Optimization (DPO)}

Fine-tuning is used to help models learn a new content without being trained on a full corpus again. More specifically, Direct Preference Optimization (DPO) is a Reinforcement Learning method used to teach models what is an accepted behavior and what is a rejected behavior \citep{rafailov_direct_2023}. According to prompts curated by humans, there is an \textit{accepted} answer and a \textit{rejected} answer. In the ChatML DPO Pairs (n=12,000). \textit{Chosen} answers are synthetically generated by GPT-4 and \textit{rejected} answers are synthetically generated by the LLaMA Model \citep{touvron_llama_2023} (see Figure~\ref{fig:bunka-dpo}). This approach is based on the assumption that GPT-4's larger number of parameters generally creates more accurate answers than LLaMA. We want to identify topics unique to GPT-4 answers which are absent in LLaMa answers. This analysis aimed to distill the unique aspects of GPT-4 performance. We hypothesized that focusing the DPO process only on prompts that lead to GPT-4 specific responses, we could accelerate the  DPO fine-tuning process.

\subsection{Method}

\begin{figure}[h] %
  \centering 
  \includegraphics[width=12cm]{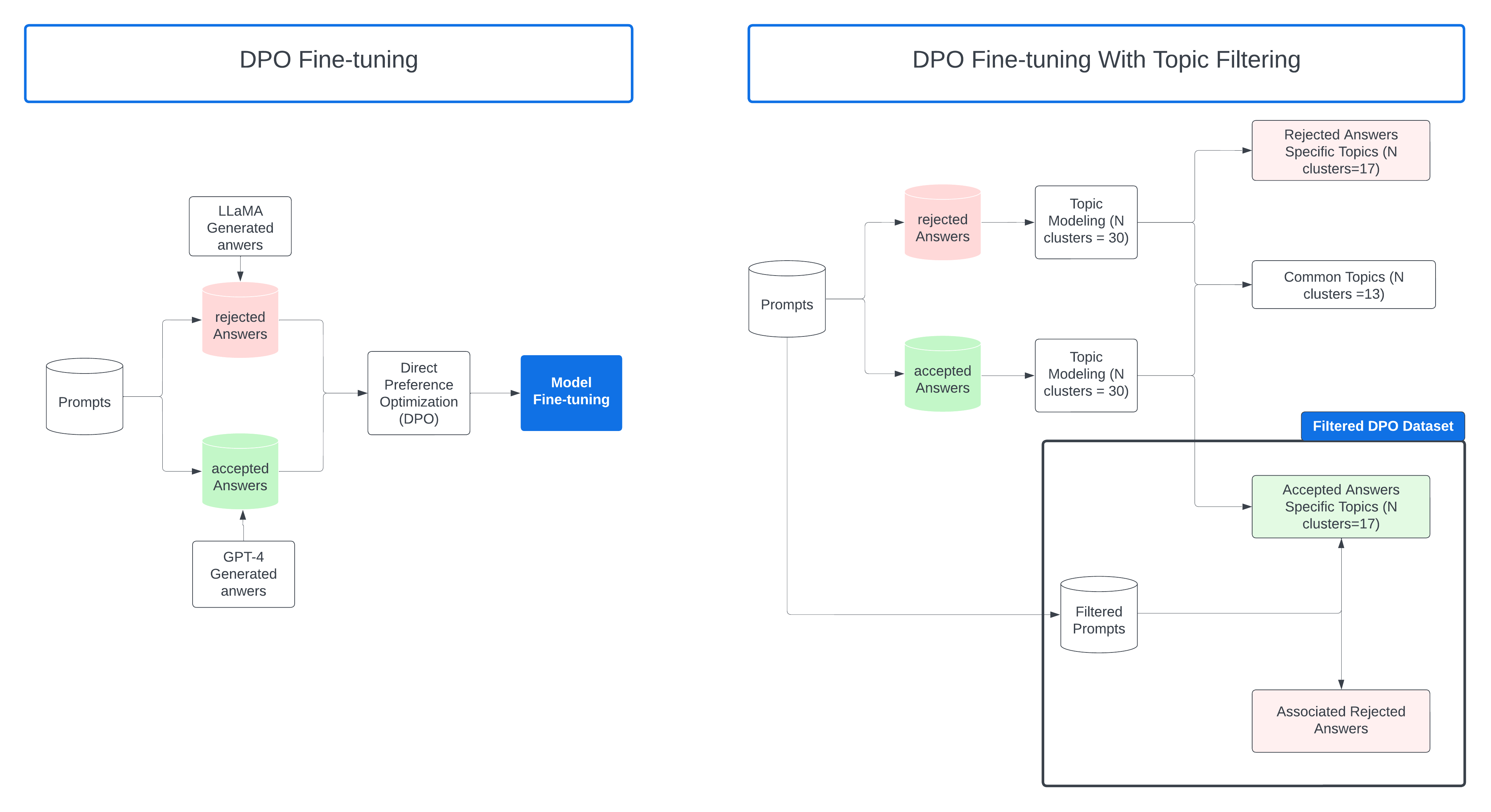} 
  \caption{Process of Direct Preference Optimisation (DPO) with and without Topic Filtering} 
  \label{fig:bunka-dpo}
\end{figure}

For each dataset (\textit{chosen} and \textit{rejected} answers), we use BunkaTopics to identify 30 topics (Given the fact we do not necessarilty need to visualize them, we arbitrarily chose 30 to capture enough granularity in our data), with each topic characterized by its 10 most specific terms. We then compared the two topic sets: we considered two topics to be overlapping if at least two of their top 10 specific terms were identical (20\% overlap). Consequently, 17 topics were found to be common between the two datasets, while 13 were distinct. We retained prompts from the \textit{chosen} answers corresponding to those 13 unique topics, reducing the data to 1/6 of the original set of \textit{prompt/accepted/rejected responses} (see Figure~\ref{fig:bunka-dpo}). The 13 topics found by BunkaTopics that are specific to GPT4 can be found in Annexe.

\begin{figure}[h] %
  \centering 
  \includegraphics[width=7cm]{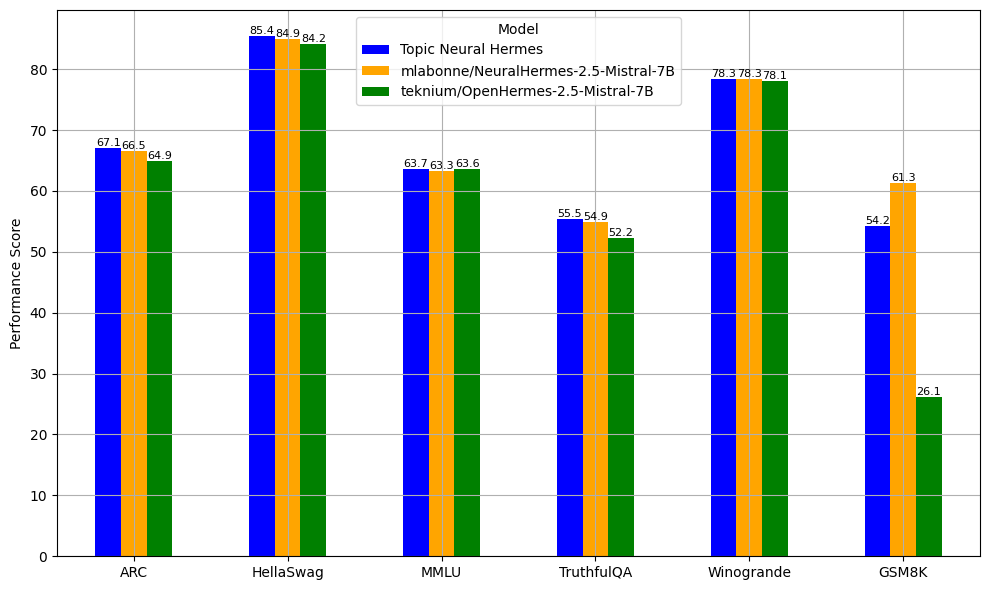} 
  \caption{Comparison of results between the DPO-filtered model (\textbf{Topic Neural Hermes}), the model fine-tuned on the full ChatML DPO Pairs (\textbf{NeuralHermes-2.5-Mistral-7B}), and the base model (\textbf{OpenHermes-2.5-Mistral-7B}). Results can be found on the HuggingFace OpenLLM leaderboard.
} 
\label{fig:comparisons}
\end{figure}

We use \textbf{OpenHermes-2.5-Mistral-7B} as a base model and fine-tune it with the filtered  \textit{prompt/accepted/rejected dataset}. We call this new model \textbf{Topic Neural Hermes}. We chose \textbf{OpenHermes-2.5-Mistral-7B} as a base model as previous work has been done to fine-tune it on the whole ChatML DPO Pairs, resulting in the \textbf{NeuralHermes-2.5-Mistral-7B} model. We could then compare our results to existing ones saving compute time. We used a A100 provided by Google Colab Pro. It took less than 30 minutes to fine-tune our model. Our findings indicate that \textbf{Topic Neural Hermes} outperforms both models across most benchmark tasks, with the exception of GSM8K (refer to Figure~\ref{fig:comparisons}).

\FloatBarrier

\section{Use Case 3: Analyzing bias using Semantic Framing Analysis}

Media coverage often frames specific subjects: they use specific vocabularies to put their subject into perspective. Kwak defines framing as the following: \textit{"Framing is a process of emphasizing a certain aspect of an issue over the others, nudging readers or listeners towards different positions on the issue even without making a biased argument.”} \citep{kwak_frameaxis_2021}. Different computational tools like FrameAxis \citep{kwak_frameaxis_2021} and OpenFraming \citep{bhatia_openframing_2021} helped study framing at scale. FrameAxis shows for instance that in the context of restaurant reviews, different frames can be used to describe a topic: \textit{inhospitable/hospitable}; \textit{active/quiet}; \textit{expensive/cheap} or \textit{unsavory/savory}. We use the concept of frames and apply them to LLMs training datasets to visually understand the relationships between a content and dedicated frames and the relationships between frames themselves to spot potential bias or imbalance which has been a core issue of LLM training \citep{gallegos_bias_2024}. 

\subsection{Framework: Semantic Frames Cartography}

\textbf{Computing Frames -} A frame is composed of two sentences or terms (like \textit{active} and textit{quiet} or \textit{expensive} and textit{cheap}). To analyze the frame of a document, we first embed the two sentences (or terms) that create the continuum of the frame. Let's denote the embedding of the first sentence as \( e_1 \), the embedding of the second sentence as \( e_2 \), and the embedding of the document as \( e_{\text{doc}} \). Following \cite{kozlowski_geometry_2019}, the frame embedding, \( e_{\text{cont}} \), is calculated as follows:
\[
e_{\text{cont}} = e_1 - e_2
\]

Then, the coordinate of the document in the Semantic Frame, denoted as \( C_{\text{Frame}} \), is computed using the cosine similarity formula between the coordinate of a document and the coordinate if the Semantic Frame:
\[
C_{\text{Frame}} = \frac{e_{\text{doc}} \cdot e_{\text{cont}}}{\|e_{\text{doc}}\| \|e_{\text{cont}}\|}
\]

We perform the same operation for the entire set of documents in our database and for a second frame, and center our plot at 0. As a result of the two frames, we obtain a 2-dimensional plot where each point represents a document and the two axes, \(x\) and \(y\), correspond to the two frames.

\begin{figure}[h] %
  \centering 
  \includegraphics[width=8cm]{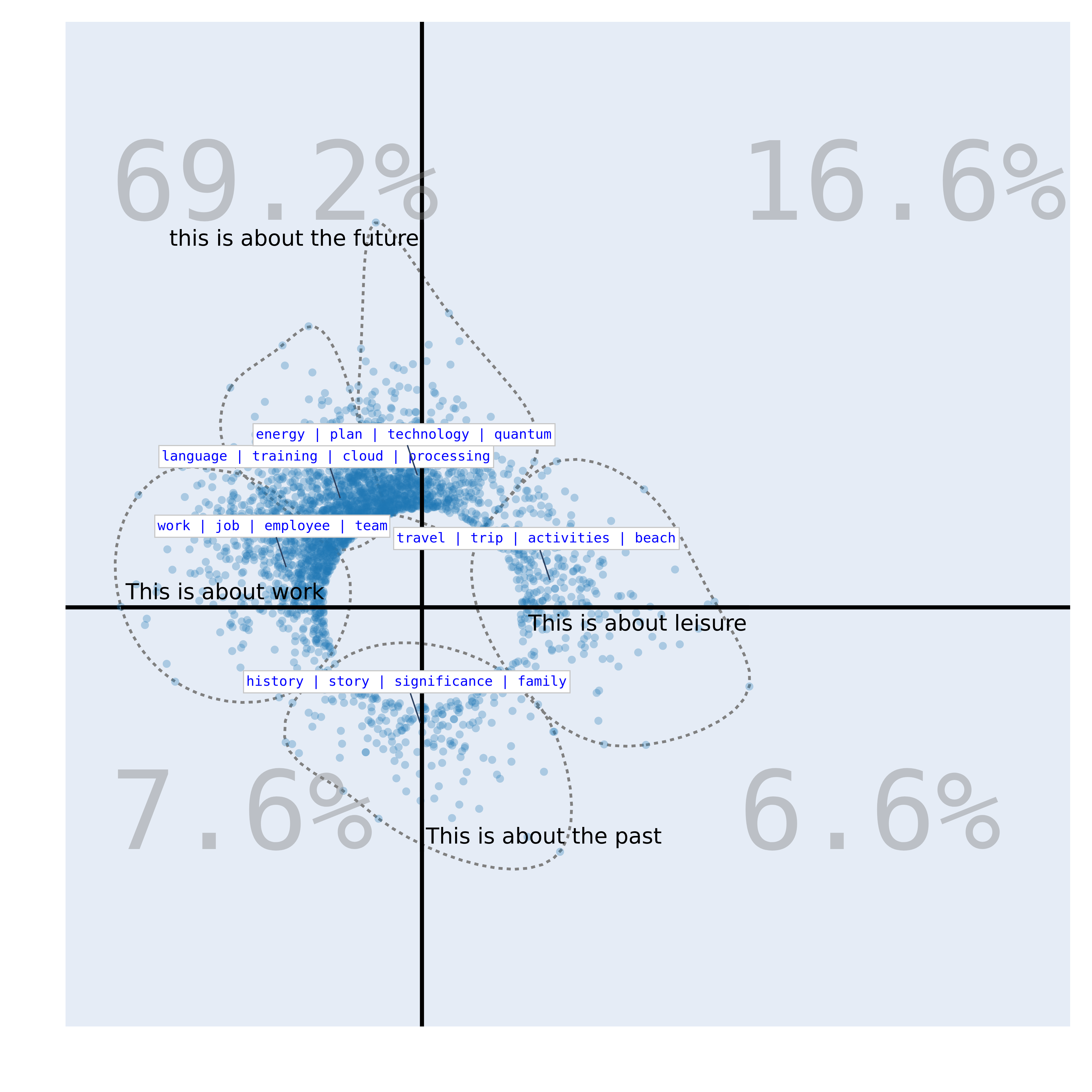} 
  \caption{Semantic Frames on \textit{past/future} and \textit{work/leisure} with Bunka Topic Modeling.} 
  \label{fig:bourdieu}
\end{figure}

In order to understand the bias of the collective-prompt dataset, Bunka uses the \textbf{UAE-Large-V1} embedding model to embed two pairs of sentences representing two semantic frames. We use \textbf{UAE-Large-V1} as it is one of the top models on the MTEB Leaderboad under 350 million parameters. We create a first semantic frame with the following sentences: \textit{“this is about the future”} and \textit{“this is about the past”} and the second semantic frame by \textit{“this is about the work”} , \textit{“this is about the leisure”}. We arbitrarily chose those two semantic frames. In order to get measures of imbalance, we calculate the percentage of documents that have values greater than 0 in both frame 1 and frame 2, values less than 0 in frame 1 and greater than 0 in frame 2, and we apply this calculation across all possible combinations (see Figure~\ref{fig:bourdieu}). Results suggest that overall there is more information about the future (85.8\%) than about the past (14.2\%) and much more work-related information (76.8\%)  than leisure-related information (13.2\%). Concepts related to \textit{Work} and \textit{Future} (69.2\%) are more related than concepts about leisure and past (6.6\%). Using BunkaTopics, we computed clusters KMeans on the new latent spaces and display the results as a map. We arbitrarily set the number of cluster to 5 to improve the visual understanding of the plot. For concepts related to \textit{Work} and \textit{Future}, we find topics  related to \textit{work-job-employee-team}, or \textit{language-training-cloud-processing}. When it comes to topics related to \textit{Future} and \textit{Leisure}, we find a topic related to \textit{travel-trip-activities-beach}. In the next part, we explain why there is a round shape in the center of the map.

\begin{figure}
    \begin{subfigure}{0.49\textwidth}
        \centering
        \includegraphics[width=\linewidth]{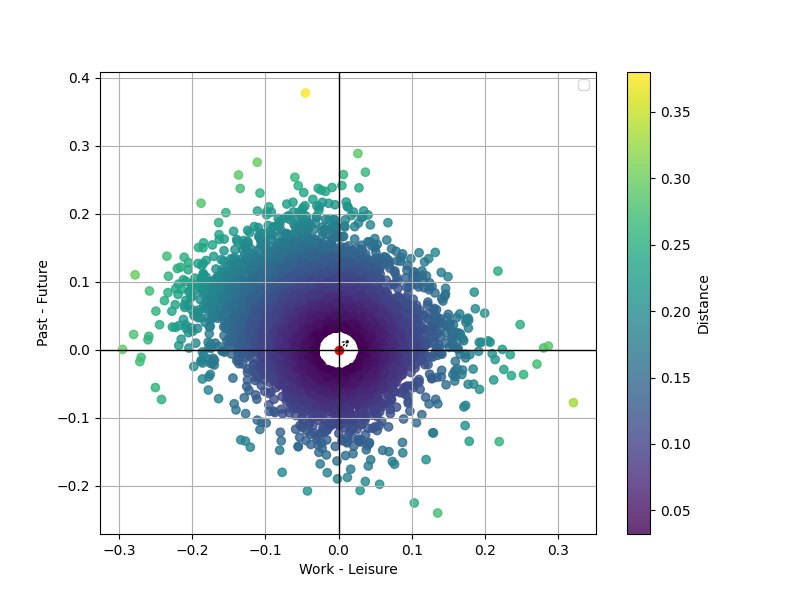}
        \caption{Filtering coefficient: 0.1}
        \label{fig:a}
    \end{subfigure}
    \hfill
    \begin{subfigure}{0.49\textwidth}
        \centering
         \includegraphics[width=\linewidth]{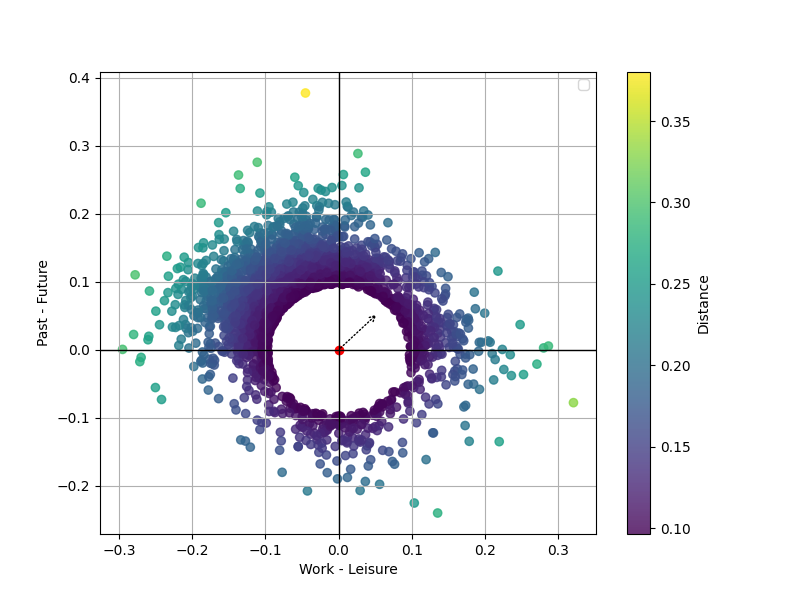}
        \caption{Filtering coefficient: 0.3}
        \label{fig:b}
    \end{subfigure}
    \caption{Bunka Semantic Framing with different levels of filtering coefficient. Every point is a document of the dataset. Distance is computed between the embeddings of the documents and the frames embeddings. There is no need to normalize given the fact that the axes are computed with the same method, but we centered around 0 for  better readability.
   } 
   \label{fig:radius}
   \end{figure}

\textbf{Filtering uncertainty -} Because the cosine similarity between embeddings only makes sense for the human mind starting from a specific threshold specific to each embedding model \citep{jose_exploration_2017} we filter out some points on the graph (see Figure~\ref{fig:radius}). The results around the center 
are the most uncertain. In order to filter out those uncertain results, we plot a circle from the center of the plot whose radius is defined with a specific coefficient. We define the coefficient from 0 to 1 and multiply it by the highest objective value either on the x-axis or the y-axis. Figure~\ref{fig:radius} is an example of the process on the axis; \textit{work-leisure} and \textit{Past-Present} of the \textbf{collective-prompt} dataset. For instance, if the point is at the center, it means that it is neither really \textit{Work}, \textit{Leisure}, \textit{Past} or \textit{Future} (see Figure~\ref{fig:radius}).

\begin{figure}
 \begin{subfigure}{0.49\textwidth}
     \centering
     \includegraphics[width=\linewidth]{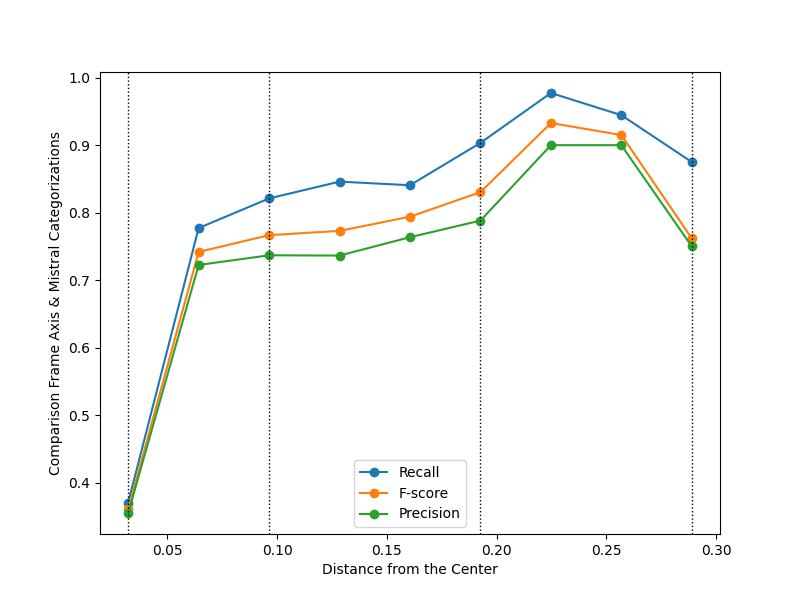}
     \caption{Future or Past}
     \label{fig:a}
 \end{subfigure}
 \hfill
 \begin{subfigure}{0.49\textwidth}
     \centering
      \includegraphics[width=\linewidth]{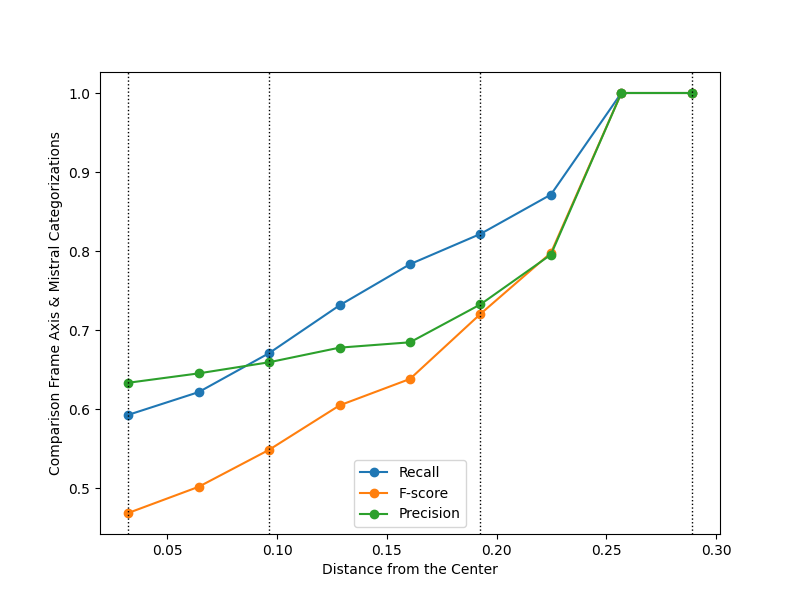}
     \caption{Work or Leisure}
     \label{fig:b}
 \end{subfigure}
 \caption{Categorization performance between Bunka Semantic Framing and zero-shot categorization using \textbf{Mistral-7B-Instruct-v0.1}  lines represent the different coefficient in order: 0.1, 0.2, 0.3, 0.4 with A. \textit{Future-Past} and B. \textit{Work-Leisure}} 
 \label{fig:classification}
\end{figure}

In our methodology, by taking the example of the frame \textit{Future - Past}, if cframe > 0, then the document is classified as the category \textit{Future}, if it <0 then the document is classified in the category \textit{Past}. For different level or radius filtering, we compare this classification methodology to the classification made by an LLM (\textbf{Mistral-7B-Instruct-v0.1}) prompted to do so (see Annexe for the prompt). This model is one of the highest performing open-source with 7 billion parameters.We show that the Frames’ classification and the LLM classification converge when filter condition by the circle radius is high enough. The convergence diminished when the number of filtered documents is too high (out of 7 remaining documents, 2 divergence leads to 70\% precision) (see Annexe). We then chose the radius (distance to the center) that gives the best results for the two frames (d=0.25) (see Figure~\ref{fig:classification}).

\FloatBarrier 

\section{Conclusion and Discussion}

We introduced Bunka, a python package using Visual Topic Modeling and Semantic Framing to analyze diverse training datasets. After explaining the architecture, we analyzed 3 Use Cases. The first use case focuses on visualizing a dataset of prompts to understand its content and comparing the results of different embedding models. We then showed that Topic Modeling can be used to filter the prompts of a Preference dataset with the idea that during the Direct Preference Optimization (DPO) process, the model will learn quickly the good answers by reducing the redundant ones (ie the prompts where \textit{chosen} and \textit{rejected} answers are too similar). For most benchmarks except GMSK8, we obtain better results with 6 times less data. Lastly, we performed Semantic Framing to analyze different types of bias in the datasets and show how we can optimize the parameters by comparing the classification results with a bigger LLM. An overall advantage of the embedding-based Semantic Framing Analysis is the cost of classification: a few seconds are needed and computation costs are limited given the small size of embedding models, while zero-shot categorization for 10,000 data using \textbf{Mistral-7B-Instruct-v0.1} took 33GB of RAM and 15 minutes to run on an A100 using vLLM \citep{kwon_efficient_2023}. 

Overall data exploration is a key component of AI transparency \& explainability, new methods like filtering by Topic Modeling are needed to get a better sense of the underlying structure of the dataset.

There are some limits to our work. The first limit is inherent to Topic Modeling where more work needs to be done regarding the right number of topics to choose and the evaluations to implement: while there is a lot of research that connect the visual aspect of data summarization and the ability for the human brain to quickly understand the content \citep{bellmund_navigating_2018}, more research should be made to understand more precisely what part of the Human Computer Interaction (HCI) really brings more sense-making and helps decision-making processes

Because there are no satisfying answers, we make it possible in our system for the user to iterate over different numbers of topics until the quality seems good enough. A solution is to systematically compare a sample within every topic with the results of a large LLM prompted to do so or to ask a human annotator to rate those topics. As topics are often domain-specific, this approach must be optimized for different industries or areas of research. Additionally, more research needs to be done to understand the relationships between chunk sizes, embedding model, reduction model, and clustering models. Every decision regarding those parameters impacts the overall results and better methods should map those decisions to the results of Topic Modeling. We show, for example, that shorter documents lead to better convergence between embedders and bigger LLM models (see Annexe). Another limit is the results in the Grade School Math 8K (GSM8K) Benchmark. While our model outperformed others on six benchmarks, it underperformed on this particular one. We hypothesize that this benchmark is content-oriented, meaning that a model's performance improves as it learns more content. This could explain why the less data the model processes, the poorer the results are. Regarding Semantic Framing, since it focuses on analyzing bias, it is important to understand the biases inherent in the embedding tool itself. Further research is necessary to effectively separate the two. Overall, the BunkaTopics package is a powerful tool for understanding and optimizing training datasets for LLMs which in turn can lead to more transparent and more efficient AI models.

\section{Code and data Availability}

Bunkatopics Package: \url{https://github.com/charlesdedampierre/BunkaTopics}

Github code: \url{https://github.com/charlesdedampierre/NeurIPS2024}

\bibliographystyle{plainnat}
\bibliography{bib}

\appendix

\renewcommand\thefigure{\thesection.\arabic{figure}}    
\section{Annexe}
\FloatBarrier 

\setcounter{figure}{0} 

\begin{figure}[ht]
 \begin{subfigure}{0.49\textwidth}
     \centering
     \includegraphics[width=\linewidth]{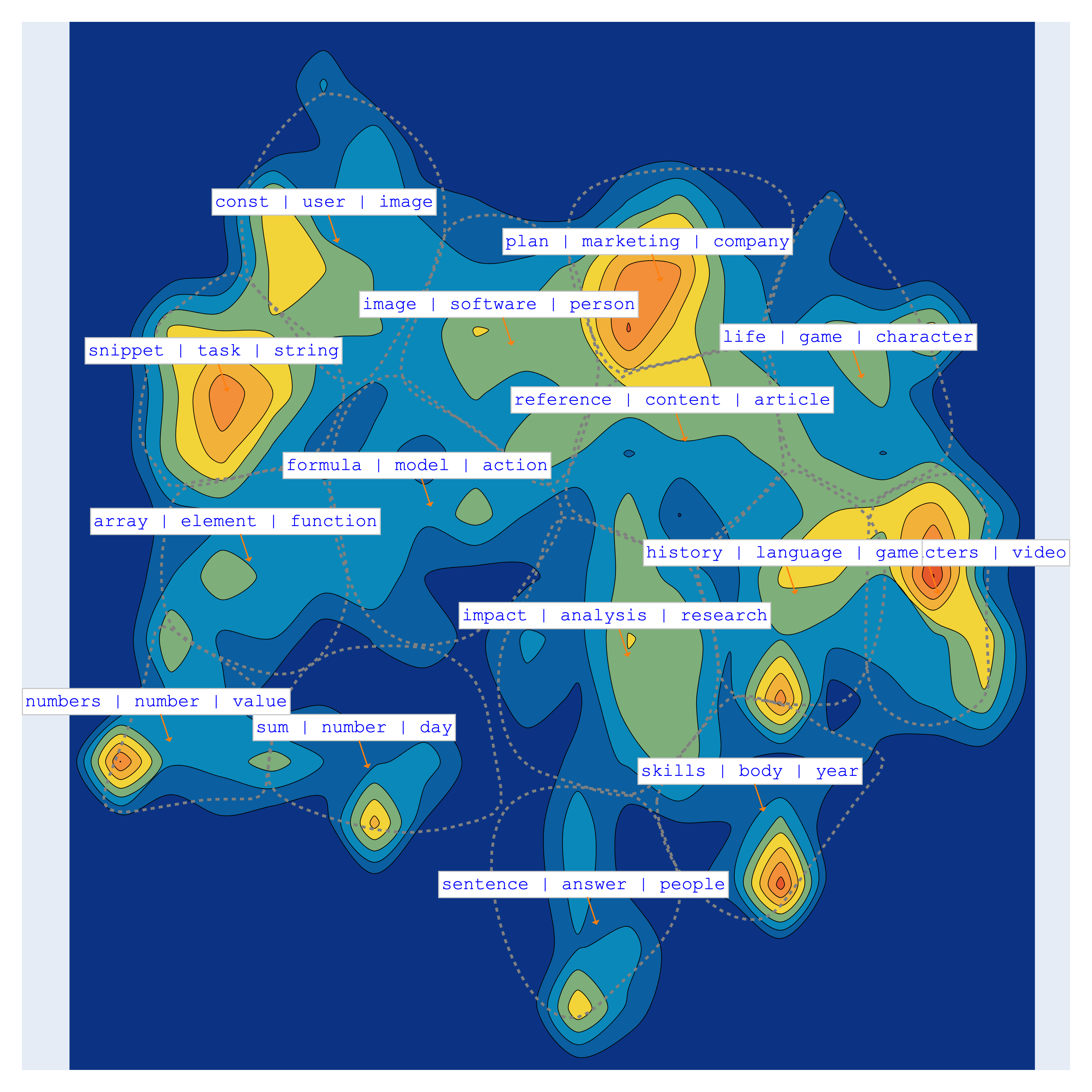}
     \caption{\textbf{all-MiniLM-L6-v2}}
     \label{fig:a}
 \end{subfigure}
 \hfill
 \begin{subfigure}{0.49\textwidth}
     \centering
     \includegraphics[width=\linewidth]{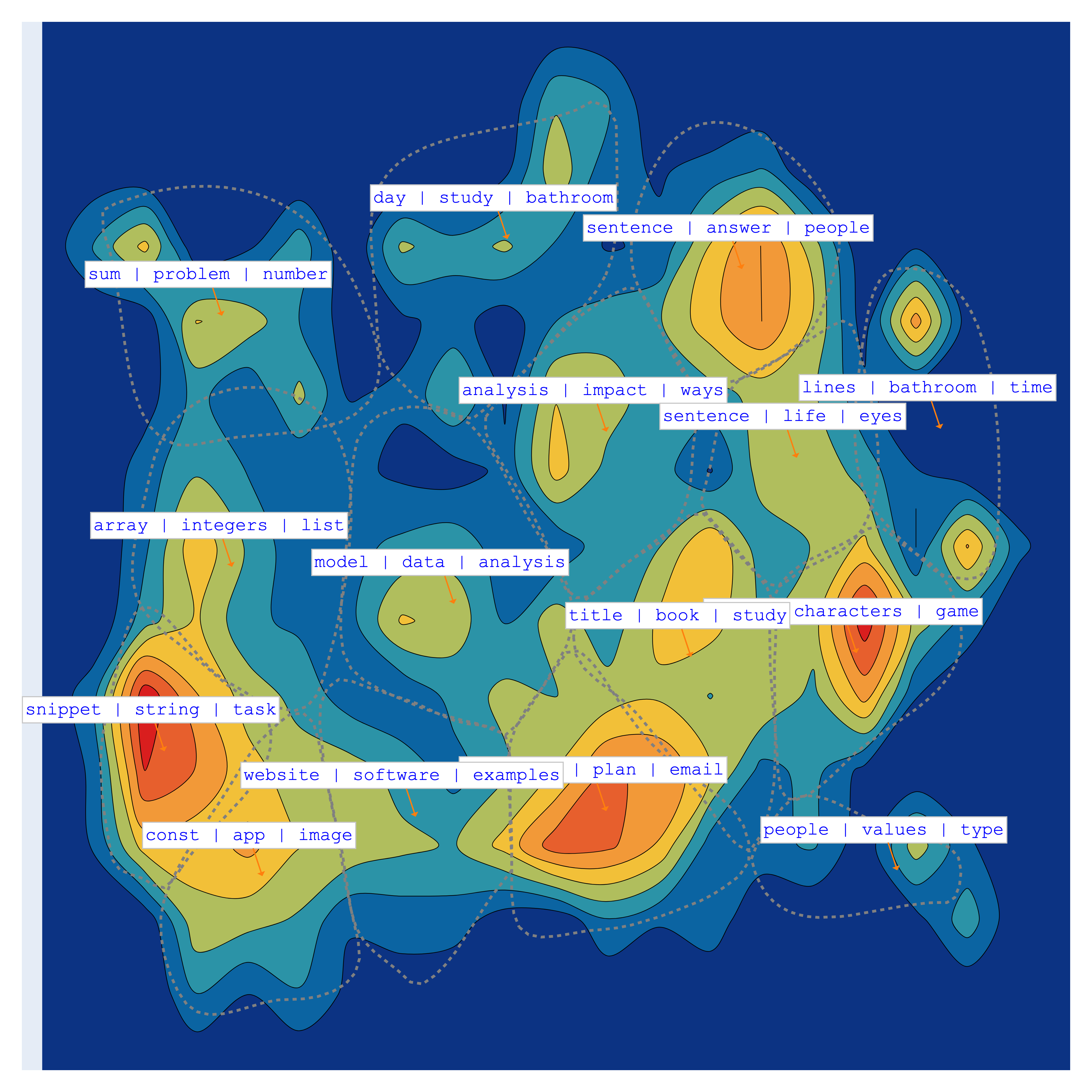}
     \caption{\textbf{bge-large-en-v1.5}}
     \label{fig:b}
 \end{subfigure}
 
 \medskip
 \begin{subfigure}{0.49\textwidth}
     \centering
     \includegraphics[width=\linewidth]{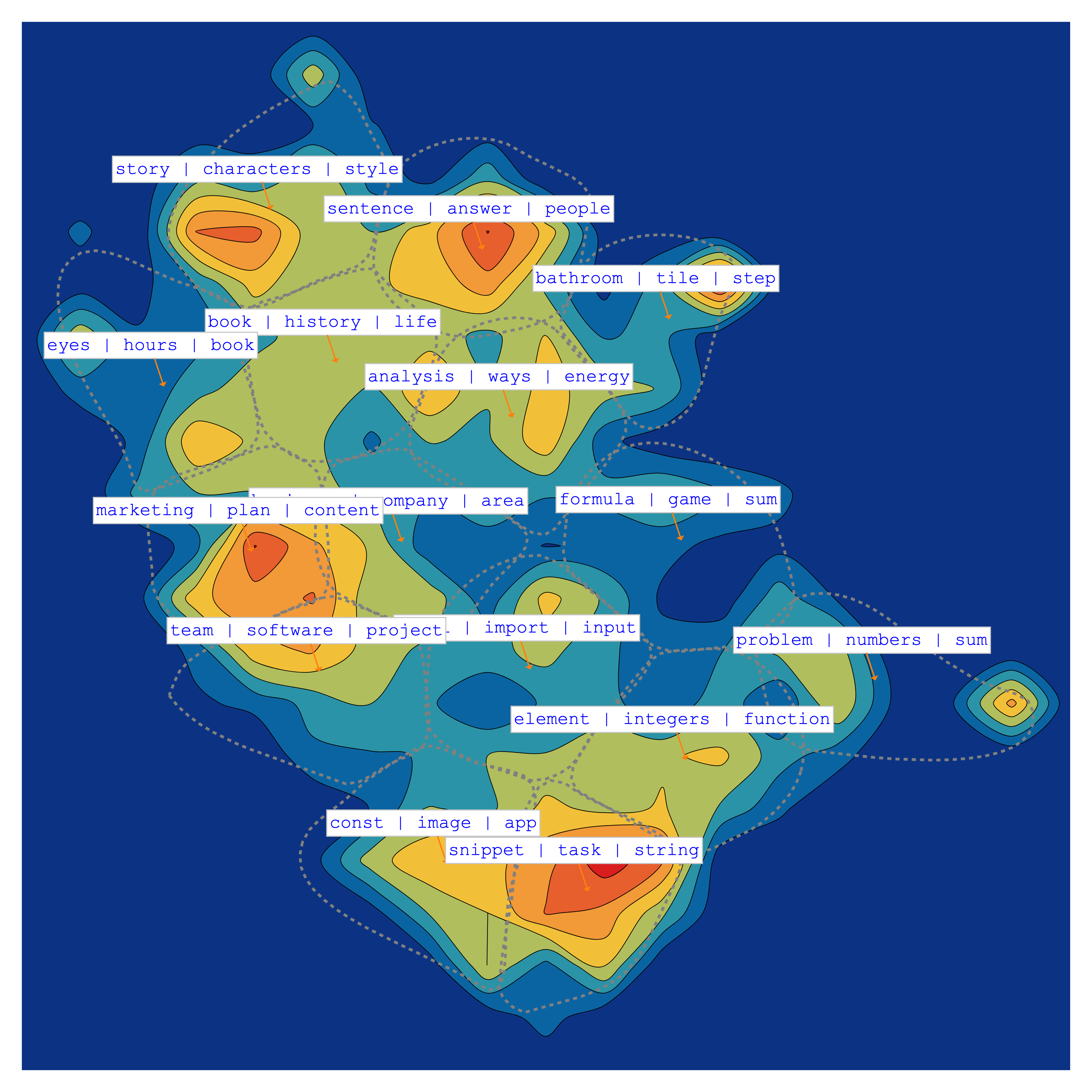}
     \caption{\textbf{UAE-Large-V1}}
     \label{fig:c}
 \end{subfigure}
 \hfill
 \begin{subfigure}{0.49\textwidth}
     \centering
     \includegraphics[width=\linewidth]{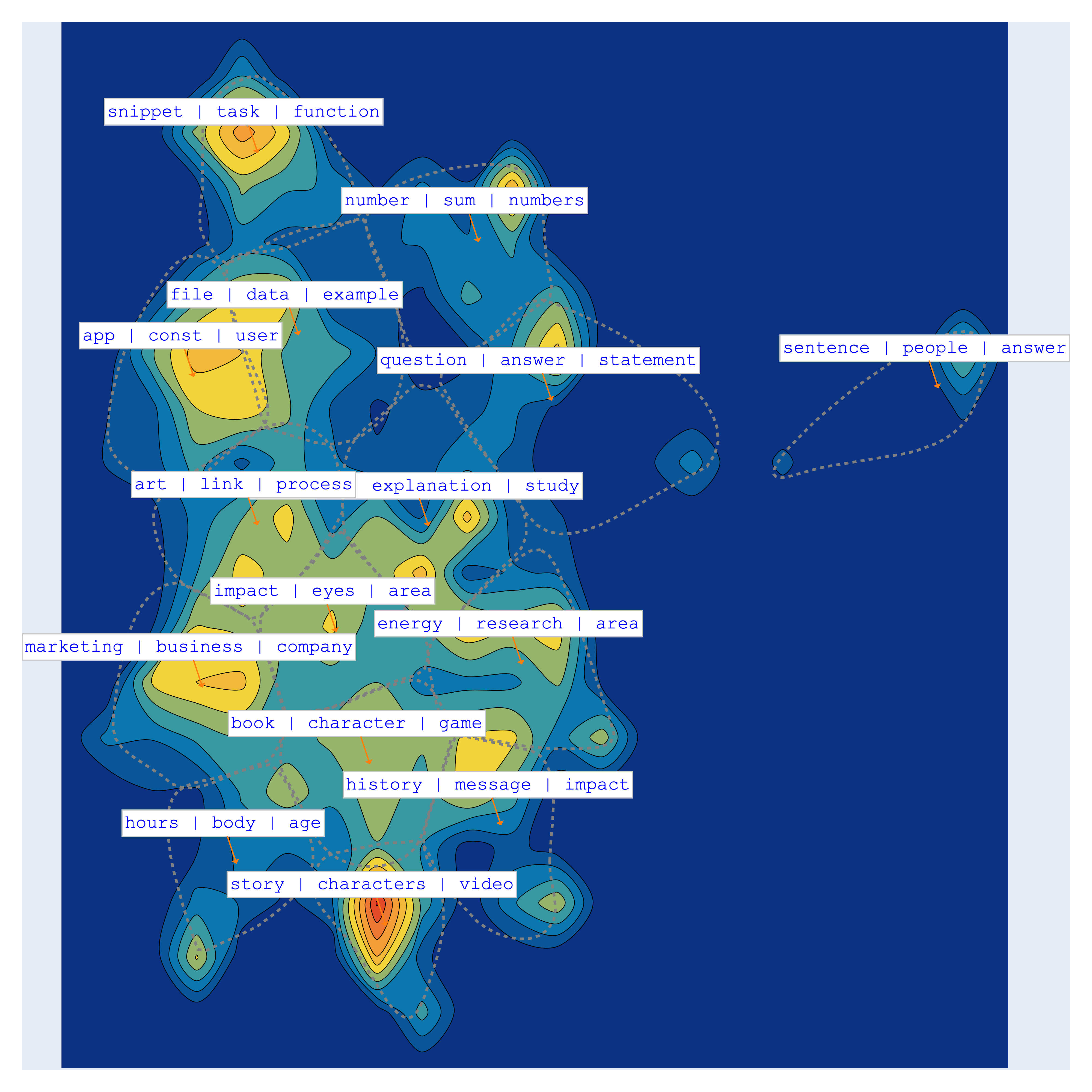}
     \caption{\textbf{multi-qa-mpnet-base-dot-v1}}
     \label{fig:d}
 \end{subfigure}
 \caption{ Topic Cartography with 15 clusters. We chose 3 specific nouns for visualization purposes. The Maps are created with 4 different embedders} 
 \label{annexe:bunka-maps-examples}
\end{figure}

\FloatBarrier 

\begin{table}[ht]
  \caption{Specific topics found by Bunnkatopics for GPT4 accepted answers}
  \label{sample-table}
  \centering
  \begin{tabular}{|m{6cm}|m{9cm}|}
    \hline
    \textbf{Topic Name}                   & \textbf{Specific Terms}                                      \\ \hline
    Emotional Dynamics                    & feelings, Quinn, Austin, minority women, teaching, schools, individual, personality, backgrounds, triggers      \\ \hline
    Global Knowledge Queries              & question, information, geography, news articles, Step, answer, capital city, pipeline system, country, analogy \\ \hline
    Digital Interactions and Queries      & questions, question, PersonX, modem, answers, effect relationship, Quora, browser, answer, e-commerce           \\ \hline
    Business and Cybersecurity            & email, businesses, initiatives, innovation, advertising papers, spam, breaches, antivirus, payments, prospects \\ \hline
    Lifestyle and Wellness                & sleep, exercise, gifts, shopping, Casey, stores, stress, headaches, options, mood                                   \\ \hline
    Wildlife Ecology                      & birds, prey, animals, species, infection, nest, eggs, bacteria, insects, kitty condo                                \\ \hline
    Environmental Science and Climate     & temperature, gases, greenhouse, emissions, perturbation, sulfur, dioxide, climate change, water, heat               \\ \hline
    Maritime and Mechanical Engineering   & ship, bowling, propulsion, beam width, Filing cabinet, LED, lane, containment area, lawnmower, rotors               \\ \hline
    Cultural and Social Dynamics         & Lindsey, museum, Kate, Rachel, Jason, Alex, Erin, conversation, Laura, exhibits                                        \\ \hline
    Political Media Analysis             & media platforms, election, politics, teenagers, elections, White House, Barack Obama, nation, Confederate, depression \\ \hline
    International Relations and Policy   & cooperation, EU, nations, alliance, NATO, European Union, member states, policy, monarch, Brexit                       \\ \hline
    Astrophysics and Physical Sciences   & electrons, km, Moon, acceleration, orbit, friction, current, asteroid, electron, collector emitter                     \\ \hline
    Film Critique and Analysis           & movie review, film, reviewer, sentiment, critic, flaws, DVD, plot, opinion, originality                                 \\ \hline
  \end{tabular}
  \label{annexe:topics-accepted}
\end{table}

\FloatBarrier 

\begin{figure}[ht]
  \medskip
  \begin{subfigure}{0.49\textwidth}
      \centering
       \includegraphics[width=\linewidth]{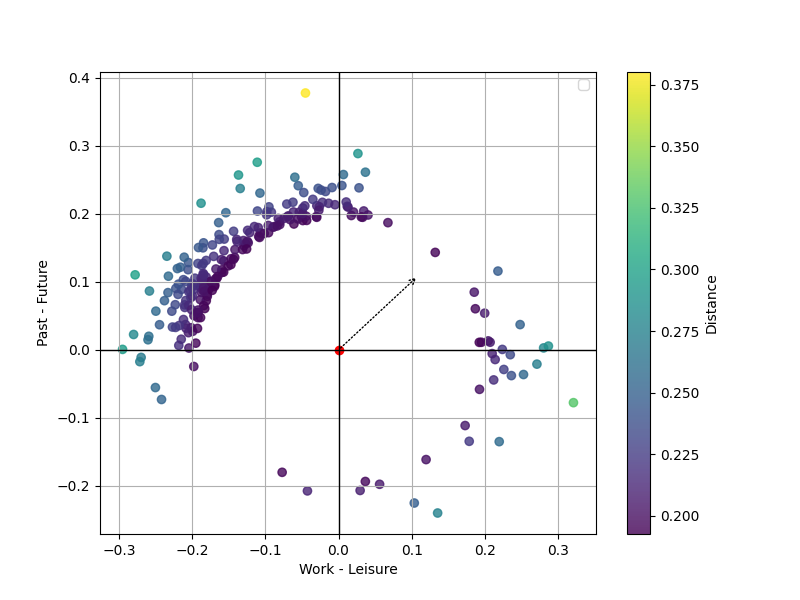}
      \caption{Filtering coefficient: 0.6}
      \label{fig:c}
  \end{subfigure}
  \hfill
  \begin{subfigure}{0.49\textwidth}
      \centering
        \includegraphics[width=\linewidth]{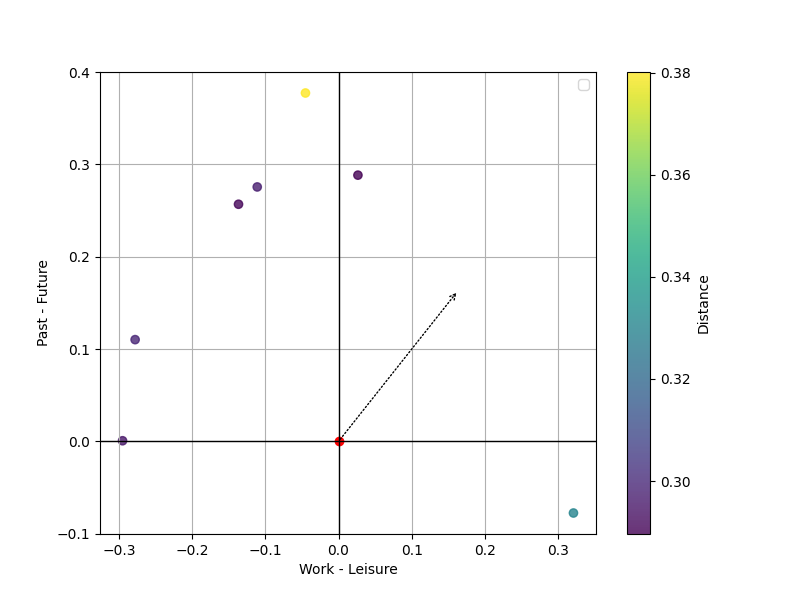}
      \caption{Filtering coefficient: 0.9}
      \label{fig:d}
  \end{subfigure}
  \caption{Bunka Semantic Framing with different levels of filtering coefficient. Every point is a document of the dataset. Axis are defined by the difference between the embeddings of their two names. Distance is computed between the embeddings of the documents and the frames embeddings. There is no need to normalize given the fact that the axes are computed with the same method, but we centered around 0 for  better readability.
 } 
 \label{annexe:radius}
 \end{figure}

 \FloatBarrier 

\begin{lstlisting}[language=Python, caption={Prompt for document classification by \textbf{mistral-7b-instruct-v0.1}}][ht]
def prompt(text):
    return f"""
    Here is a sentence:

    {text}

    Choose 2 categories: one from these three: "leisure", "work", or "None", AND one from these three: "future", "past", or "None".

    Give the result as a JSON:

    {{'category_1': 'result_1', 'category_2': 'result_2'}}

    Answer:
    """
\end{lstlisting}

\FloatBarrier 

\begin{figure}[ht]
 \centering
 \begin{subfigure}{0.49\textwidth}
     \centering
     \includegraphics[width=\linewidth]{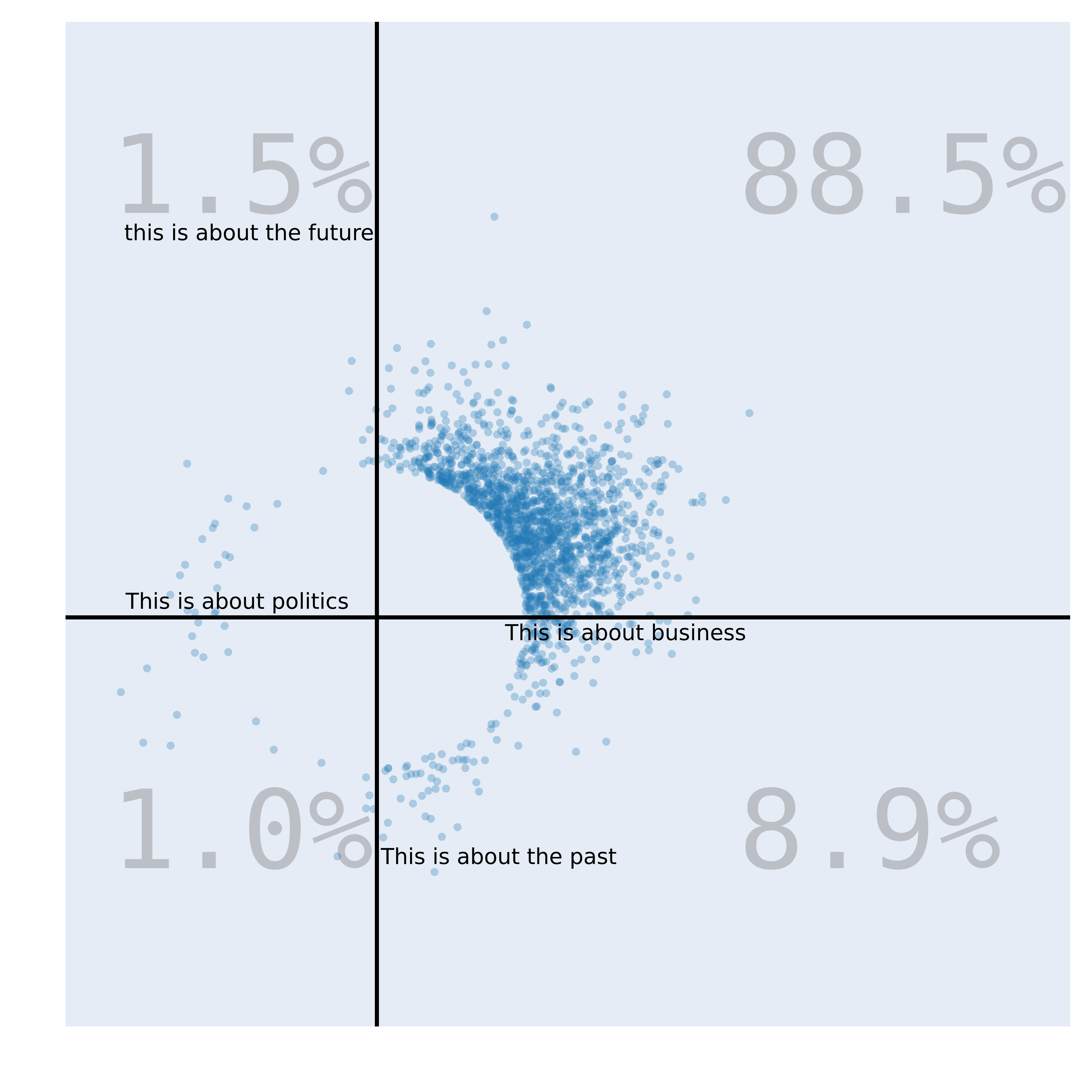}
     \caption{\textit{politics/busines} \& \textit{past/future}}
     \label{fig:a}
 \end{subfigure}
 \hfill
 \begin{subfigure}{0.49\textwidth}
     \centering
     \includegraphics[width=\linewidth]{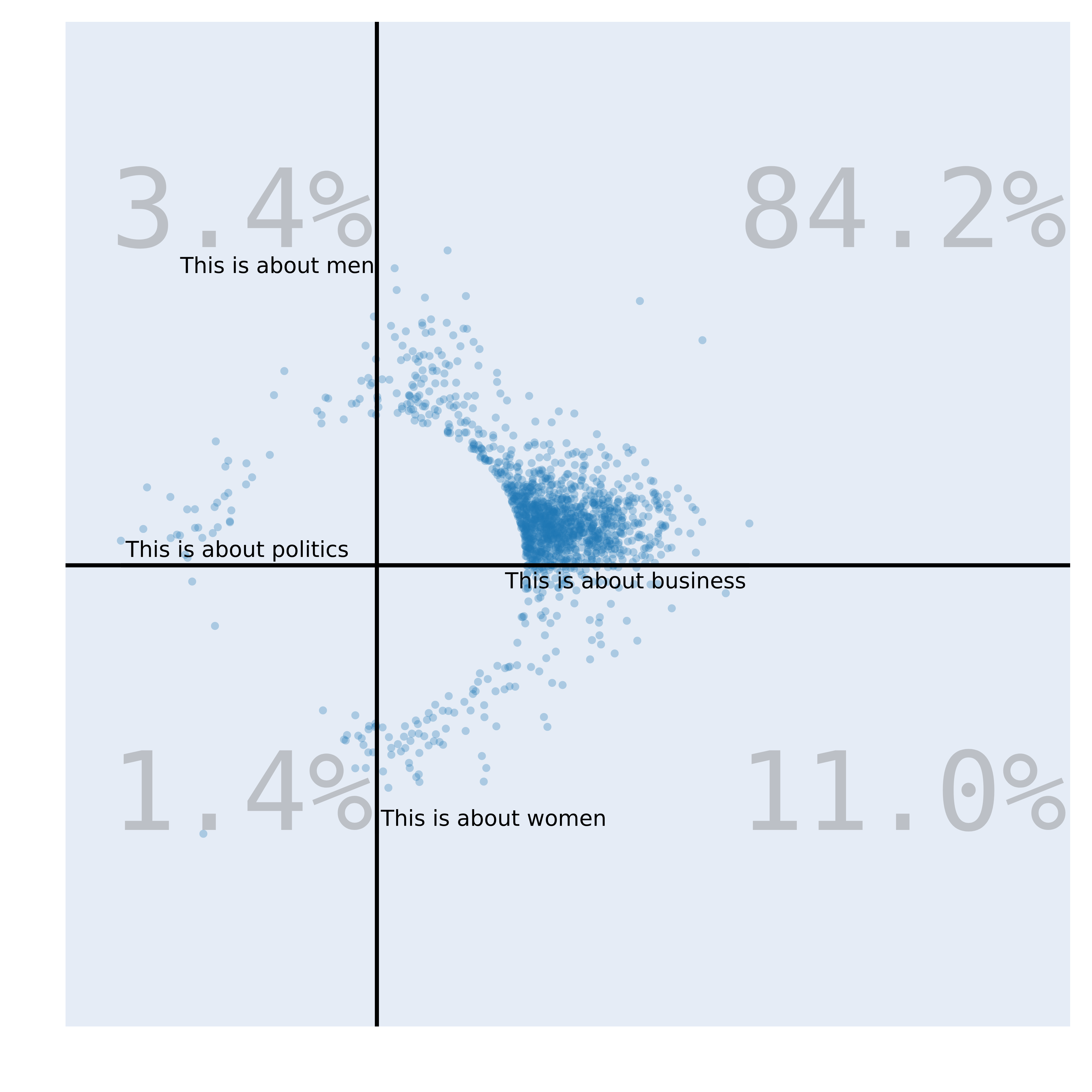}
     \caption{\textit{politics/business} \& \textit{men/women}}
     \label{fig:b}
 \end{subfigure}
 
 \medskip 
 \begin{subfigure}{0.49\textwidth}
     \centering
     \includegraphics[width=\linewidth]{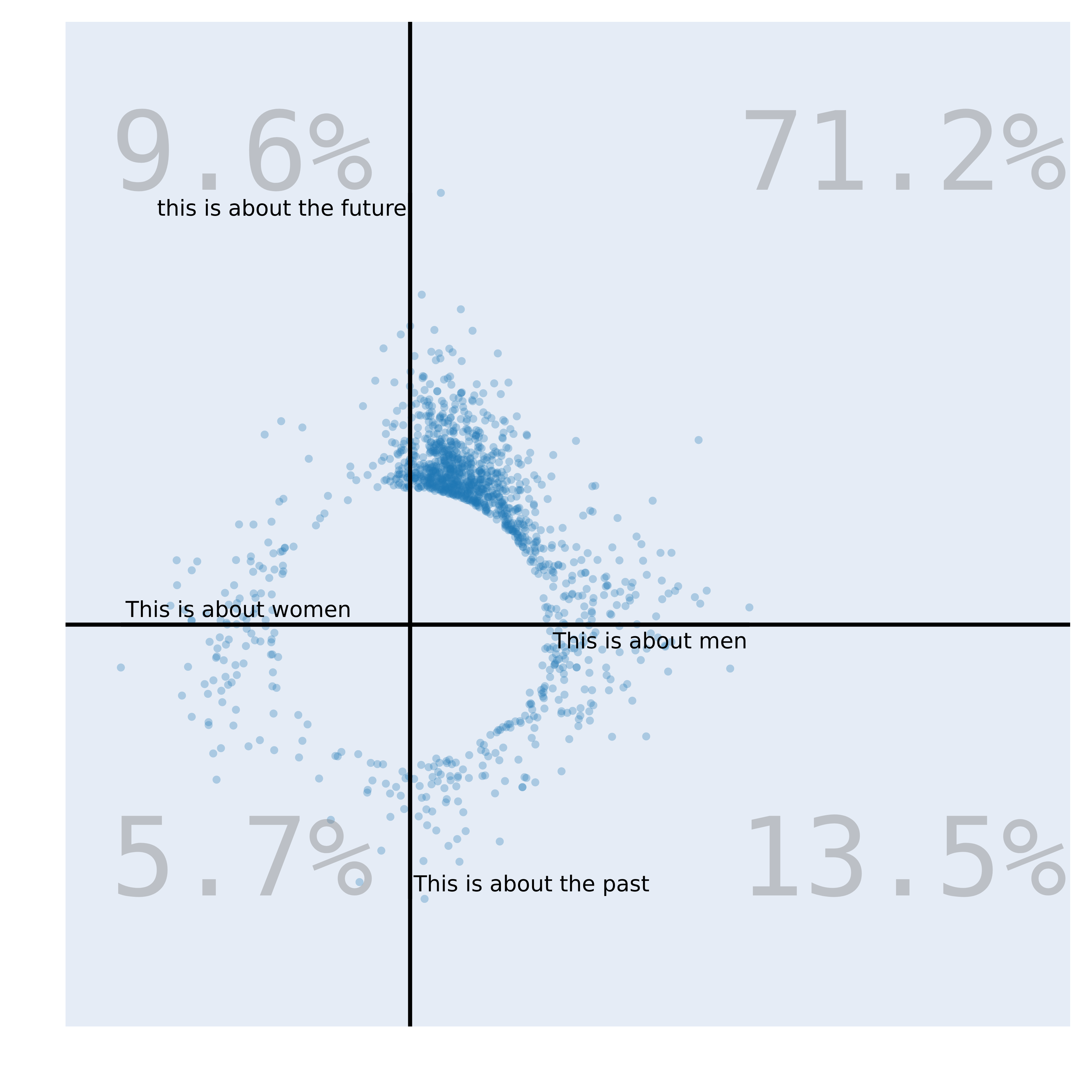}
     \caption{\textit{men/women} \& \textit{past/future}}
     \label{fig:c}
 \end{subfigure}
 \hfill
 \begin{subfigure}{0.49\textwidth}
     \centering
     \includegraphics[width=\linewidth]{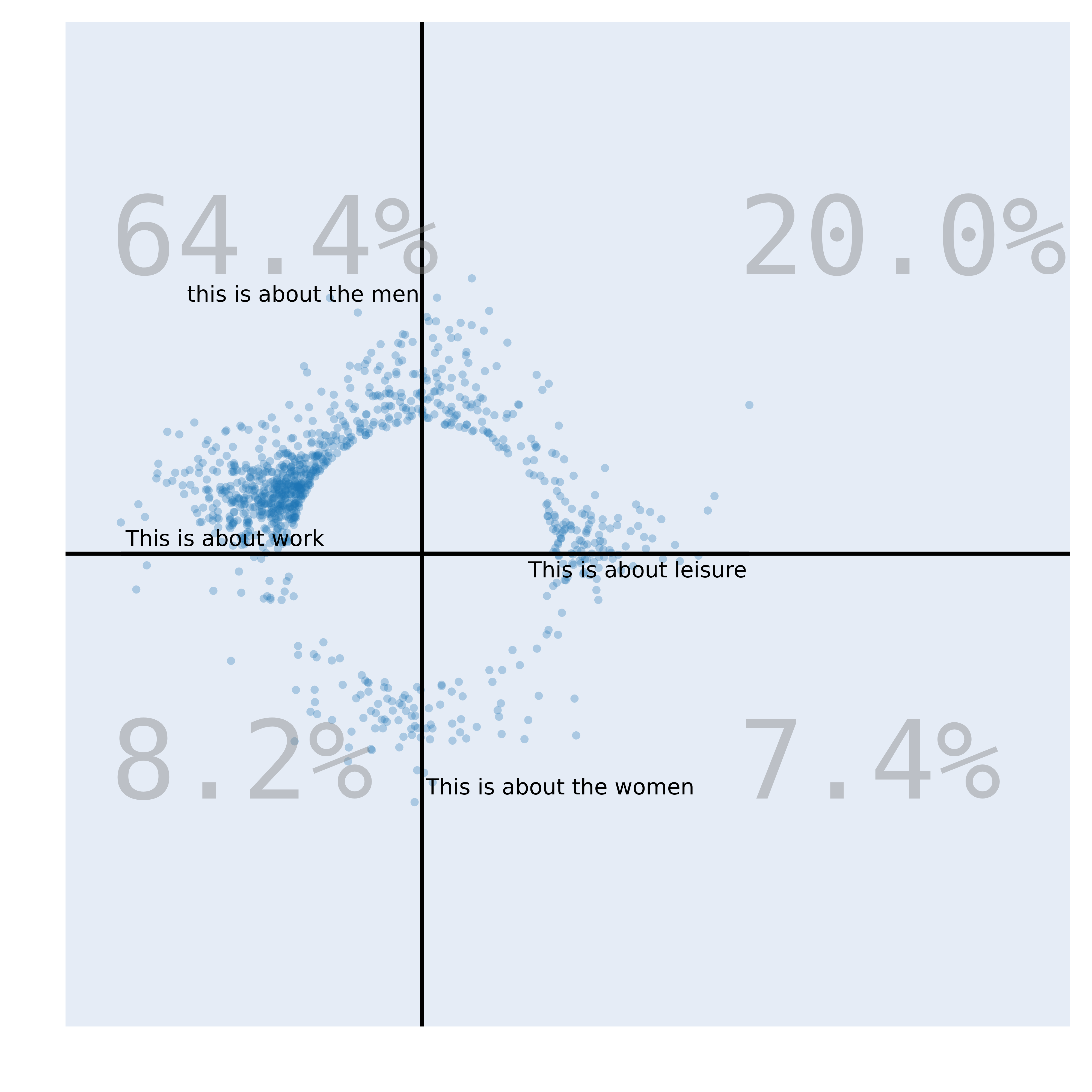}
     \caption{\textit{men/women} \& \textit{work/leisure}}
     \label{fig:d}
 \end{subfigure}
 \caption{Semantic framing using different sentences. Our results suggest that the dataset is biased towards the concepts of \textit{future}, \textit{business}, \textit{men} and \textit{work} as compared to their opposite concepts. Also, the concept of \textit{men} is more likely to be associated with the concept of the \textit{future} than the concept of \textit{women}. The concept of \textit{women} is more likely to be associated with the concept of \textit{leisure} than with the concept of \textit{work} in comparison with the concept of \textit{men}. }
 \label{annexe:bourdieu-examples}
\end{figure}

\begin{figure}[ht]
 \begin{subfigure}{0.49\textwidth}
     \centering
     \includegraphics[width=\linewidth]{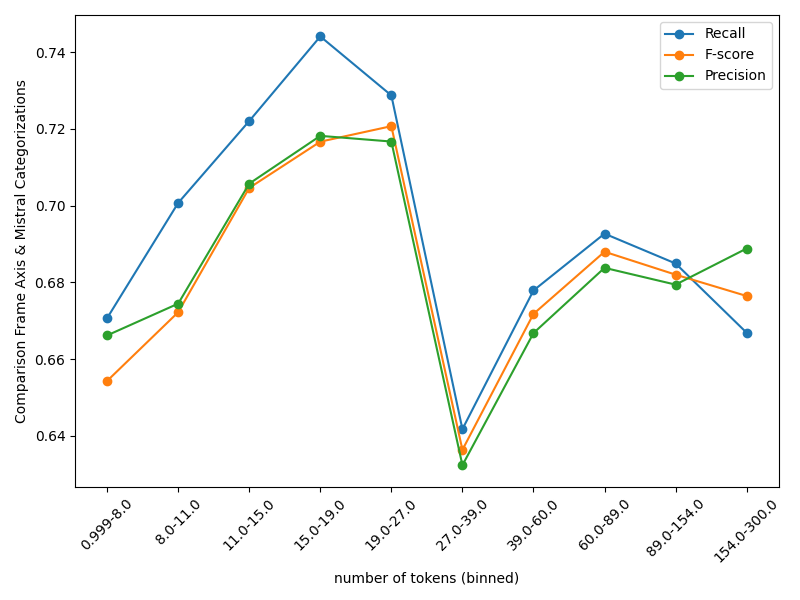}
     \caption{\textit{Future or Past}}
     \label{fig:a}
 \end{subfigure}
 \hfill
 \begin{subfigure}{0.49\textwidth}
     \centering
      \includegraphics[width=\linewidth]{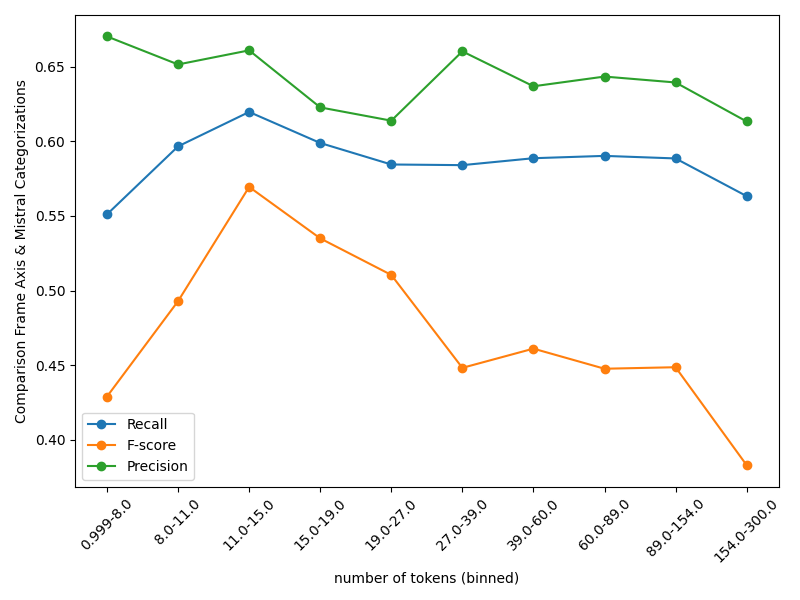}
     \caption{\textit{Work or Leisure}}
     \label{fig:b}
 \end{subfigure}
 \caption{Categorization performance between Bunka Semantic Framing and zero-shot categorization using \textbf{Mistral-7B-Instruct-v0.1} based on the length of tokens. We also note that the length of tokens has an impact on the classification results: there seems to be an optimum between 10 and 30 tokens where the Semantic Framing model and the \textbf{Mistral-7B-Instruct-v0.1} converge towards the same answer. We see a drop for the \textit{Future-Past} categorization drops in the range [27-39] tokens.}
\end{figure}

\end{document}